# Reliability of Language-Model Judges for Quantitative Scientific Question Answering: Evidence from Perovskite Photovoltaics

Qiwei Han[*1,2], Chi Zhou[2] and Gaoxiang Yuan[3]

[1]Department of Chemistry, Duke University, Durham, NC 27708, USA

[2]Department of Computer Science, Georgia Institute of Technology, Atlanta, GA 30332, USA

[3]Department of Electrical and Computer Engineering, Tandon School of Engineering, New York University, Brooklyn, NY 11201, USA

[*]Corresponding author.

**Abstract**

Evaluating retrieval-augmented language models increasingly relies on automated model judges, yet their reliability on quantitative scientific synthesis remains largely unmeasured. Here, we evaluate automated grading on a quantitative benchmark for perovskite photovoltaics, where all 755 ground-truth answers are deterministically computed from 45,441 experimental device records. We evaluate a frontier judge (Claude Opus 5.5) alongside two local open-weight judges (Qwen3-14B and Qwen3-8B) across 4,530 free-text answers from six retrieval architectures under three regimes: reference-augmented (with the computed key), evidence-augmented (with retrieved passages and query logs), and unassisted (question and answer alone). Equipped with the reference key, Claude Opus reproduces ground truth near-perfectly (Cohen's $\kappa = 0.982 \pm 0.003$) and preserves exact architectural rankings ($\tau = 1.00$). Withholding the reference causes evaluation fidelity to collapse: frontier agreement drops to $\kappa = 0.557 \pm 0.022$, and providing retrieved literature passages yields virtually no improvement ($\kappa = 0.576 \pm 0.024$). Without reference keys, local judges degrade to near-random agreement ($\kappa = 0.087$ and $0.112$), inverting the leaderboard to rank a hallucinating closed-book baseline first ($\tau \leq 0.20$). Furthermore, SQL-augmented generations induce severe evaluative blind spots: absent a reference, the frontier judge accepts 46.6% of erroneous SQL answers, overestimating architecture accuracy as 66.4% against its true 52.6%. These findings demonstrate that model-based evaluation in the physical sciences cannot rely on unstructured text inspection; rigorous benchmarking requires reference ground truth precomputed from structured measurement records.

## 1 Introduction

Language models increasingly serve as literature discovery assistants across the physical sciences, with retrieval-augmented generation (RAG) providing the dominant paradigm for grounding model generations in published evidence [1–3]. Because model outputs take the form of open-ended natural language prose, evaluation has broadly shifted from manual grading to automated "LLM-as-a-judge" architectures that inspect candidate responses and assign correctness verdicts. While model-based judges align moderately with human preferences on conversational tasks [4] and form the core scoring engine of popular RAG evaluation frameworks [5, 6], these evaluators are themselves almost universally calibrated against subjective human ratings or proxy preferences. Recent studies demonstrate that model judges exhibit severe operational pathologies, including length bias [4], sensitivity to the order in which answers are presented [7], and an inability to perform robust multi-step numerical verification [8].

In quantitative empirical disciplines such as materials science and photovoltaics, these evaluative vulnerabilities trigger an intractable circular validation dilemma. Scientific inquiries rarely concern isolated qualitative facts; rather, they demand cross-literature numerical synthesis—identifying which published studies exceed an efficiency threshold, computing distribution medians across fabrication cohorts, or tracking material performance trends across decades. Evaluating such claims introduces an acute operational trap: manual expert re-annotation is physically impossible at literature scale, as no human reviewer can aggregate hundreds of device records by hand to verify a candidate statistic, forcing practitioners to rely on automated LLM judges. Yet existing evaluation paradigms suffer from a fatal scale mismatch: standard frameworks attempt reference-free grading by supplying the judge with retrieved text snippets, but a handful of retrieved passages (typically up to eight passages spanning 2,400 words) cannot mathematically verify an aggregate property derived from dozens or hundreds of independent articles. As a consequence, reported accuracies across the scientific RAG literature currently rest on evaluators whose own fidelity on quantitative domain synthesis remains completely unmeasured.

Here, we resolve this validation dilemma by introducing an experimental methodology where ground truth is mathematically absolute rather than subjectively authored. Utilizing a dual-representation literature corpus encompassing 45,441 experimental device records from the perovskite solar cell literature [9–12] (pairing full-text articles with relational database tables), we use a quantitative benchmark of 755 template-derived questions, built in our previous study [10], whose answers and supporting citation sets are deterministically computed via relational queries. This deterministic baseline allows us to treat the language-model judge as an empirical subject under laboratory conditions, directly measuring its calibration against absolute ground truth. We evaluate 4,530 free-text answers generated across six distinct retrieval architectures by an open 14B model, testing three automated judges across three controlled information regimes: (1) *reference-augmented grading*, where the judge receives the candidate answer alongside the precomputed numerical key and tolerance rule; (2) *evidence-augmented grading*, where the judge receives the retrieved literature passages and SQL execution logs without the ground-truth key; and (3) *unassisted grading*, evaluating the candidate answer against the question text alone. We evaluate a frontier model (Claude Opus 5.5, deployed as independent grading agents) alongside two local open-weight judges (Qwen3-14B and Qwen3-8B) served on a single consumer GPU.

Our findings reveal that access to an authoritative reference key—rather than access to retrieved text passages—governs whether automated judges can be trusted. Equipped with the precomputed reference, Claude Opus reproduces computed ground truth almost perfectly ($\kappa = 0.982 \pm 0.003$) and preserves exact architectural rankings ($\tau = 1.00$). Withholding the reference key causes immediate evaluative collapse: frontier agreement drops to $\kappa = 0.557$, and providing retrieved literature passages yields negligible relief ($\kappa = 0.576$), confirming the scale mismatch between snippet retrieval and corpus-level synthesis. For local open-weight judges, reference-free evaluation produces an inverted leaderboard: both Qwen3-14B and Qwen3-8B erroneously rank a zero-shot closed-book baseline as the top-performing system ($\tau = 0.07$ and $0.20$), falsely accepting 67.9% and 91.3% of its errors. Crucially, structured SQL augmentation exposes an acute evaluative vulnerability: without an authoritative reference key, Claude Opus accepts 46.6% of erroneous database-derived claims, skewing the architecture's perceived accuracy upward to 66.4% against its true 52.6%. These results demonstrate that in empirical physical sciences, plausibility cannot substitute for truth: reliable evaluation of scientific literature assistants fundamentally requires reference ground truth precomputed from structured measurement records.

# 2 Experimental Design and Benchmark Construction

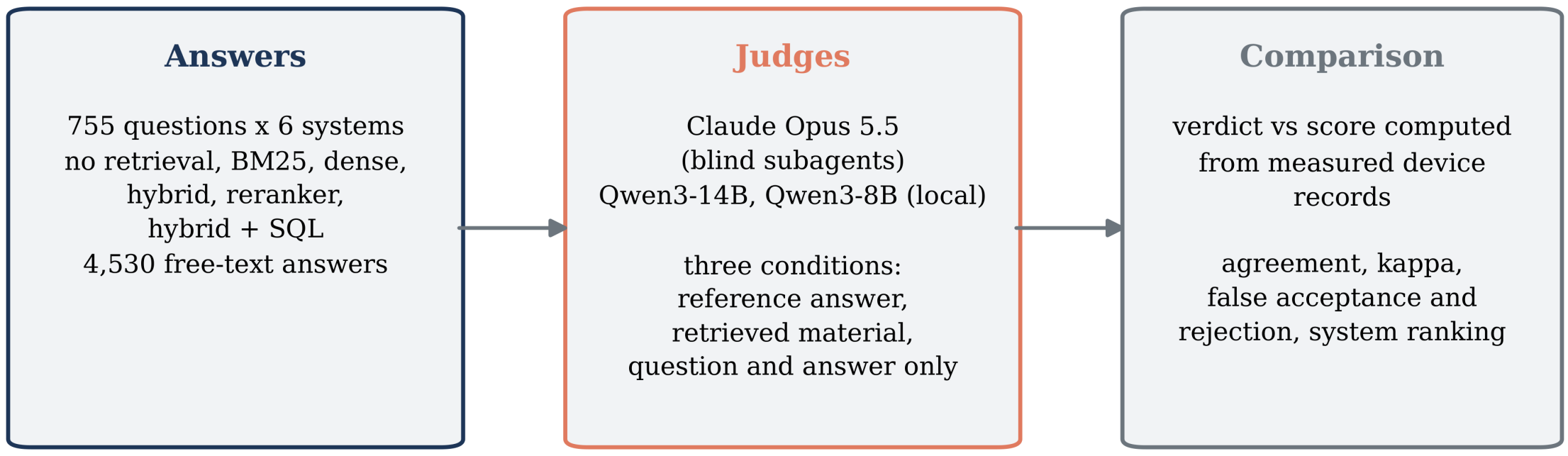

Figure 1: **Overview of benchmark evaluation methodology.** Candidate answers (4,530 natural-language generations across 755 quantitative questions from six retrieval architectures) are evaluated by three language-model judges under three information conditions: reference answer, retrieved material, and question and answer only. Judge verdicts are benchmarked directly against deterministic ground truth computed from 45,441 experimental photovoltaic records.

## 2.1 Benchmark Questions and Generative Systems

We utilize the quantitative question-answering benchmark introduced in Ref. [10] (Fig. 1), comprising 755 template-derived queries partitioned into historical literature through 2021 (371 questions derived from curated device records) and prospective literature from 2022 to 2026 (384 questions derived from experimental device records extracted and verified from full-text papers). The queries span seven distinct task categories: single-paper parameter lookups, filtered candidate lists, maxima, paper counts, distribution medians, pairwise material comparisons, and multi-year temporal trends.

We evaluate six retrieval architectures answering via an open Qwen3-14B model: zero-shot closed-book generation, sparse keyword retrieval (BM25), dense semantic retrieval, hybrid retrieval, hybrid retrieval with cross-encoder reranking, and hybrid retrieval augmented with read-only SQL execution over the device database. For every query, each system generated a natural-language response (two to five sentences articulating the empirical claim and citations) terminating in the structured target value. Across all 4,530 system generations, the pooled empirical accuracy across the six evaluated retrieval architectures is 20.4% under deterministic scoring rules: within 1% relative error (with at least a 0.05 percentage point floor) for lookups; within 1% relative error with matching citation DOI for maxima; within 5% relative error for medians; exact matching for counts and candidate DOI sets; and consistent directionality with magnitude agreement within 25% for comparisons and trends.

## 2.2 Judge Models and Information Regimes

We evaluate automated grading across three distinct information regimes without disclosing system identity:

1. *Reference-augmented grading* (reference answer and tolerance rule), where the judge receives the candidate answer alongside the precomputed ground-truth key and tolerance rule;
2. *Evidence-augmented grading* (retrieved material and query logs), where the judge receives the retrieved text passages (up to eight passages per query) and, for the SQL system, the executed queries and their results, without the ground-truth key; and

3. *Unassisted grading* (question and candidate answer alone), where the judge evaluates the candidate answer solely against the question text.

Evaluations compare a frontier model against two local open-weight models. The frontier evaluator, Claude Opus 5.5, was deployed as autonomous grading agents initialized with a fixed instruction file. Answers were labelled with random keys that hide the system that produced them; batches were filled in system order, so the first batches of each condition mix the no-retrieval and BM25 answers and every later batch holds the answers of one system. Batch sizes were set to 25 items for reference and unassisted conditions, and 10 items for the evidence condition to accommodate context limits. Judges evaluated the natural-language claim of each candidate answer (with trailing structured JSON keys stripped to prevent formatting cues) and assigned a binary verdict (`correct` or `incorrect`) via standardized JSON output. Each agent started from an empty context and graded either one batch (for the first 1,510 answers of the unassisted condition) or four consecutive batches (otherwise), writing each batch's verdicts before reading the next; when an agent stopped early, its incomplete batches were graded again by a fresh agent. Local judges (4-bit quantized Qwen3-14B and Qwen3-8B) [13] were served via SGLang [14] on a single NVIDIA RTX 3090 GPU (24 GB) with greedy decoding (temperature $T = 0$). Re-grading audits over 300 randomly sampled items confirmed high verdict reproducibility: test-retest consistency reached 99.7% under reference augmentation, 97.7% with evidence passages, and 97.3% under unassisted evaluation, with aggregate agreement shifting by less than 1.0 percentage point. Batch context scaling similarly showed negligible variance: multi-round batched grading preserved 96.0% verdict concordance relative to single-batch evaluation (across 1,508 evaluated answers), with aggregate ground-truth agreement shifting by only 1.1 percentage points.

### 2.3 Evaluation Metrics and Statistical Calibration

Judge performance is evaluated against deterministic benchmark ground truth using five primary metrics: raw accuracy agreement, chance-corrected Cohen's $\varkappa$ [15], false acceptance rate (percentage of incorrect generations rated correct), false rejection rate (percentage of correct generations rated incorrect), and aggregate reported accuracy. To evaluate architecture selection fidelity, we compute Kendall's rank correlation coefficient $\tau$ [16] between reported system accuracies and empirical benchmark performance. All statistical intervals represent 95% nonparametric bootstrap confidence intervals over 10,000 question resamples (where resampling a question bundles answers across all six architectures), reported as estimate $\pm$ bootstrap standard error (SE).

## 3 Results and Discussion

### 3.1 Agreement with Computed Ground Truth

Evaluating judge verdicts against computed benchmark ground truth reveals that access to the reference answer key dictates evaluation fidelity (Fig. 2 and Table 1). Under reference augmentation, Claude Opus reproduces computed ground truth on 99.4% of all 4,530 generations ($\varkappa = 0.982 \pm 0.003$), exhibiting a false acceptance rate of merely 0.1 $\pm$ 0.1% and a false rejection rate of 2.3 $\pm$ 0.5%. Consequently, its aggregate reported accuracy (20.1 $\pm$ 0.9%) aligns directly with true benchmark accuracy (20.4%). Under identical reference conditions, Qwen3-14B reaches $\varkappa = 0.673 \pm 0.019$: it rejects a larger share of correct answers (16.6 $\pm$ 1.6%) than it accepts of wrong ones (10.3 $\pm$ 0.8%), but because wrong answers far outnumber correct ones, its reported accuracy (25.2 $\pm$ 1.0%) still exceeds the true 20.4%. In contrast, the smaller Qwen3-8B judge shifts toward the opposite error pattern, exhibiting greater leniency by accepting 17.0 $\pm$ 0.8% of wrong answers while rejecting only 7.3 $\pm$ 1.1% of correct answers under the reference ($\varkappa = 0.621 \pm 0.018$, inflating reported accuracy to 32.5 $\pm$ 1.0%).

Withholding the reference answer key causes catastrophic evaluation collapse that dwarfs architectural differences between models. Without a reference, Claude Opus agreement drops to $\varkappa = 0.557 \pm 0.022$, falsely rejecting 30.7 $\pm$ 2.1% of correct answers due to an inability to independently verify corpus statistics. Crucially, providing retrieved literature passages does not remediate this deficit: frontier agreement remains stagnant at $\varkappa = 0.576 \pm 0.024$ (false rejection 33.2 $\pm$ 2.3%). Because eight retrieved passages provide negligible coverage over distribution medians, extrema, or multi-paper cohorts, evidence passages cannot equip a judge to verify literature-scale aggregates. For open-weight local evaluators, reference-free grading degrades to near-chance performance across model sizes: Qwen3-14B drops to $\varkappa = 0.087 \pm 0.016$ (accepting 45.7 $\pm$ 1.2% of errors; 48.4% reported accuracy), while Qwen3-8B reaches $\varkappa = 0.112 \pm 0.013$ (accepting 58.0 $\pm$ 1.0% of errors; 62.1% reported accuracy), both overstating true corpus accuracy by more than twofold. Providing evidence passages raises their agreement from 55.2% to 67.7% (Qwen3-14B) and from 49.2% to 70.5% (Qwen3-8B), and more than doubles their $\varkappa$ (to 0.226 $\pm$ 0.023 and 0.248 $\pm$ 0.024), yet evaluation fidelity remains poor. Neither the 8B nor the 14B open-weight model can reliably evaluate quantitative generations without reference verification: absent precomputed keys, both local models suffer from high error rates, failing to distinguish correct from incorrect answers.

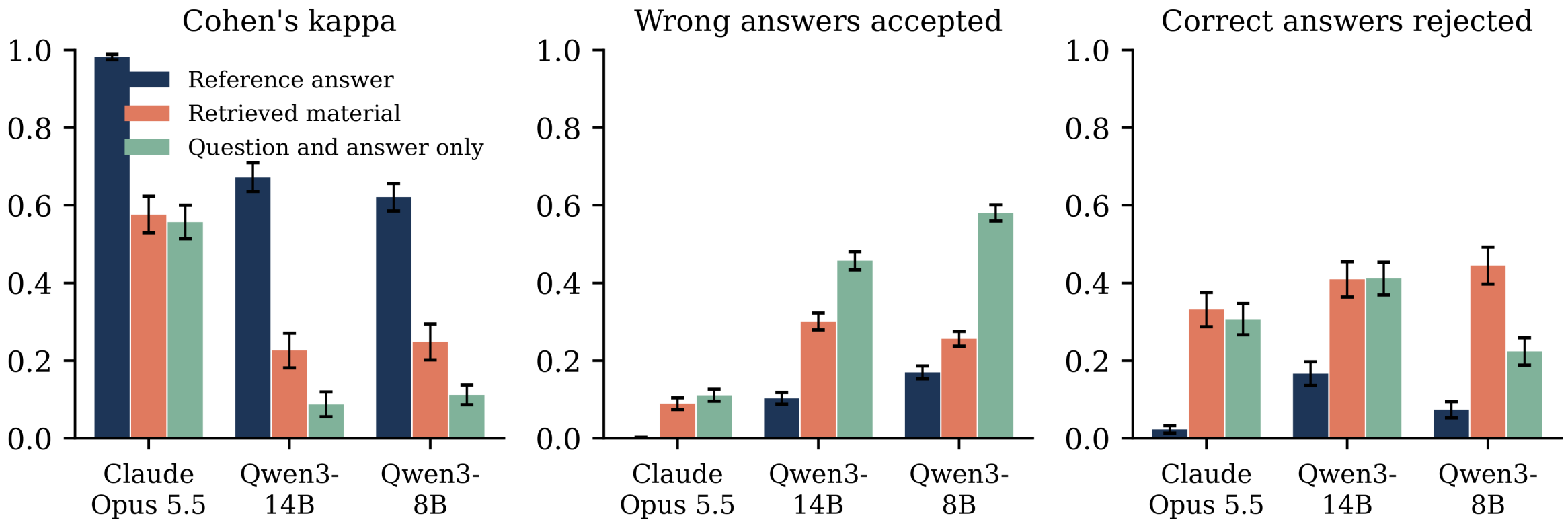

Figure 2: **Agreement of the judges with computed ground truth.** Cohen's $\varkappa$ (left), wrong answers accepted (false acceptance rate, centre), and correct answers rejected (false rejection rate, right) for each judge and evaluation condition, pooled across all 4,530 system generations. Error bars represent 95% bootstrap confidence intervals across questions; numerical values are reported in Table 1.

Table 1: **Verdicts against computed ground truth, pooled across retrieval systems.** Values are estimate $\pm$ bootstrap standard error (SE) over 4,530 answers (755 questions); the pooled empirical accuracy across the six evaluated architectures is 20.4%.

| Judge | Condition | Agreement (%) | $\varkappa$ | Wrong accepted (%) | Correct rejected (%) | Reported accuracy (%) |
|---|---|---|---|---|---|---|
| Claude Opus 5.5 | Reference answer | 99.4 ± 0.1 | 0.982 ± 0.003 | 0.1 ± 0.1 | 2.3 ± 0.5 | 20.1 ± 0.9 |
| | Retrieved material | 86.1 ± 0.8 | 0.576 ± 0.024 | 8.9 ± 0.8 | 33.2 ± 2.3 | 20.8 ± 1.0 |
| | Question and answer | 84.9 ± 0.7 | 0.557 ± 0.022 | 11.1 ± 0.8 | 30.7 ± 2.1 | 23.0 ± 0.9 |
| Qwen3-14B | Reference answer | 88.4 ± 0.7 | 0.673 ± 0.019 | 10.3 ± 0.8 | 16.6 ± 1.6 | 25.2 ± 1.0 |
| | Retrieved material | 67.7 ± 1.0 | 0.226 ± 0.023 | 30.1 ± 1.1 | 40.9 ± 2.3 | 36.0 ± 1.1 |
| | Question and answer | 55.2 ± 1.0 | 0.087 ± 0.016 | 45.7 ± 1.2 | 41.1 ± 2.1 | 48.4 ± 1.1 |
| Qwen3-8B | Reference answer | 85.0 ± 0.7 | 0.621 ± 0.018 | 17.0 ± 0.8 | 7.3 ± 1.1 | 32.5 ± 1.0 |
| | Retrieved material | 70.5 ± 0.9 | 0.248 ± 0.024 | 25.6 ± 1.0 | 44.5 ± 2.4 | 31.7 ± 1.0 |
| | Question and answer | 49.2 ± 1.0 | 0.112 ± 0.013 | 58.0 ± 1.0 | 22.4 ± 1.8 | 62.1 ± 1.0 |

### 3.2 Architectural Ranking and the Inverted Leaderboard

In comparative RAG benchmarks, the primary utility of an automated judge lies in system selection—correctly ranking architectures by empirical capability (Fig. 3). Under reference augmentation, Claude Opus establishes perfect ranking fidelity ($\tau = 1.00$), ordering all six systems without inversion. Absent reference keys, frontier ranking correlation drops to $\tau = 0.87$ with evidence passages and $\tau = 0.73$ unassisted. Importantly, this ranking correlation ($\tau = 0.87$) is anchored by the wide performance gulf separating the extreme endpoints (SQL at 52.6% versus zero-shot at 4.2%): among competing text retrievers where adjacent true accuracy gaps are 1.3, 4.0, and 0.7 percentage points (BM25 at 19.6%, hybrid at 18.3%, reranking at 14.3%, and dense retrieval at 13.6%), evidence-augmented Claude exhibits a single adjacent swap between dense and reranked retrieval (13.1% versus 12.8% reported, separated by 0.7 points empirically). Furthermore, the apparent alignment of frontier evidence-based grading with true aggregate performance is a dangerous mathematical mirage. While reporting an overall accuracy of 20.8 ± 1.0% (superficially close to the true 20.4%), this macroscopic score arises solely through mutual cancellation of severe item-level errors: the judge simultaneously rejects nearly one-third (33.2%) of correct scientific answers while falsely accepting 8.9% of outright errors. Such error masking also affects individual architectures: on the SQL-augmented architecture, evidence-augmented grading exhibits spurious calibration by reporting a deceptively aligned 52.2% accuracy against the true 52.6%, masking a 26.0% false acceptance rate through mutual error cancellation (with 24.2% false rejection). In the unassisted regime, this failure turns into gross inflation, overestimating SQL accuracy at 66.4% (falsely accepting 46.6% of errors). Practitioners evaluating RAG pipelines at the aggregate level are thus misled into trusting evaluators whose per-query verdicts are fundamentally uncalibrated.

For local open-weight judges, reference-free evaluation generates an inverted leaderboard. While Qwen3-14B recovers coarse rank ordering under reference augmentation ($\tau = 0.73$), its ranking capability collapses to $\tau = 0.33$ with retrieved passages and $\tau = 0.07$ unassisted. Under unassisted evaluation, Qwen3-14B ranks the closed-book zero-shot model (empirical accuracy 4.2%) as the single best architecture across the benchmark, falsely accepting 67.9% of its erroneous generations. Qwen3-8B exhibits identical pathology, accepting 91.3% of closed-book errors and placing zero-shot generation first ($\tau = 0.20$). Deploying reference-free local judges would lead practitioners to draw disastrous architectural conclusions—erroneously favoring unretrieved closed-book generation over retrieval-augmented pipelines.

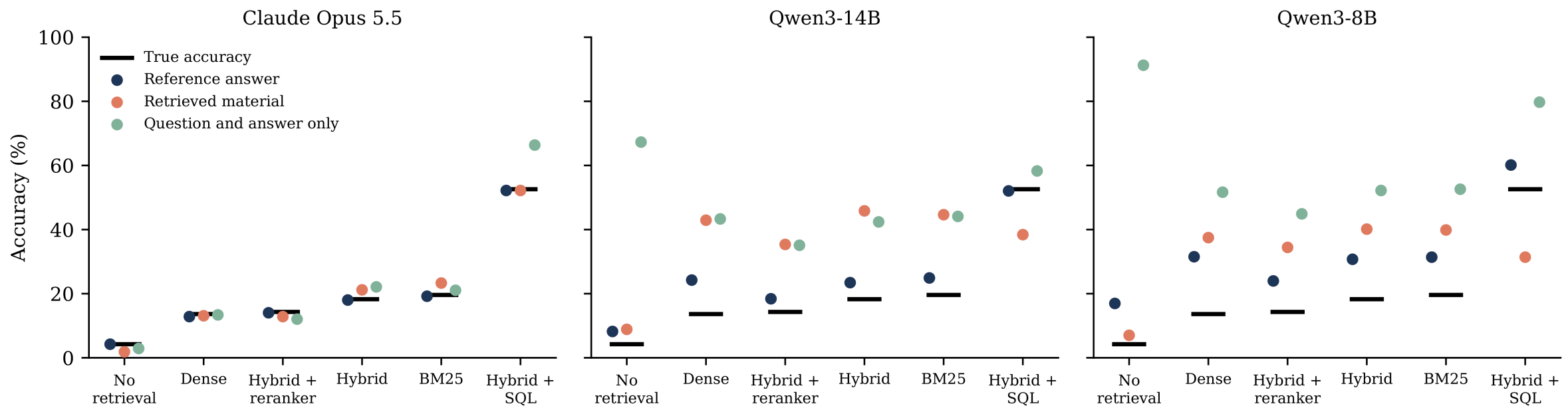

Figure 3: **Accuracy reported by each judge across retrieval systems.** Systems are ordered along the $x$-axis by true benchmark accuracy (horizontal black bars). Colored points display reported accuracy under each evaluation condition (navy: reference answer; coral: retrieved material; green: question and answer only).

### 3.3 Structured Query Vulnerability and the SQL Deception

Augmenting language models with structured SQL execution over relational tables yields the benchmark's highest performance, yet simultaneously produces the most deceptive failure modes for automated evaluators (Fig. 4). In the unassisted regime, Claude Opus falsely ac-cepts 46.6% of incorrect SQL-derived answers, overestimating architecture accuracy at 66.4% against its true 52.6%. In contrast, the same judge accepts only 6% to 11% of erroneous outputs from text-based RAG baselines. Natural-language summaries derived from relational queries prove particularly challenging for reference-free evaluators, which struggle to detect errors in structured computational statements.

Providing judges with executed SQL queries and database output logs reduces false acceptance on SQL answers from 46.6% to 26.0%, enabling partial computational auditing. Deliberation traces reveal that access to query syntax allowed Claude Opus to detect subtle programmatic bugs: computing arithmetic means instead of medians, employing loose LIKE filters that conflated lead-free tin cells with tin–lead alloys, or writing composition filters that counted triple-cation formulations as two-cation systems. However, execution trace inspection remains deeply inadequate for benchmark evaluation: even when supplied the full query text and its results (up to 50 rows), the frontier judge still fails to detect more than one in four erroneous claims (26.0% false acceptance). Only supplying the precomputed reference answer key eliminates this vulnerability, suppressing false acceptance to 0.8%. Local evaluators prove even more defenseless against structured deception: unassisted false acceptance reaches 53.9% for Qwen3-14B and 72.3% for Qwen3-8B, and remains high (26.8% for Qwen3-14B, while Qwen3-8B's 19.0% false acceptance reflects indiscriminate rejection, discarding 57.4% of correct SQL answers) even when provided execution traces.

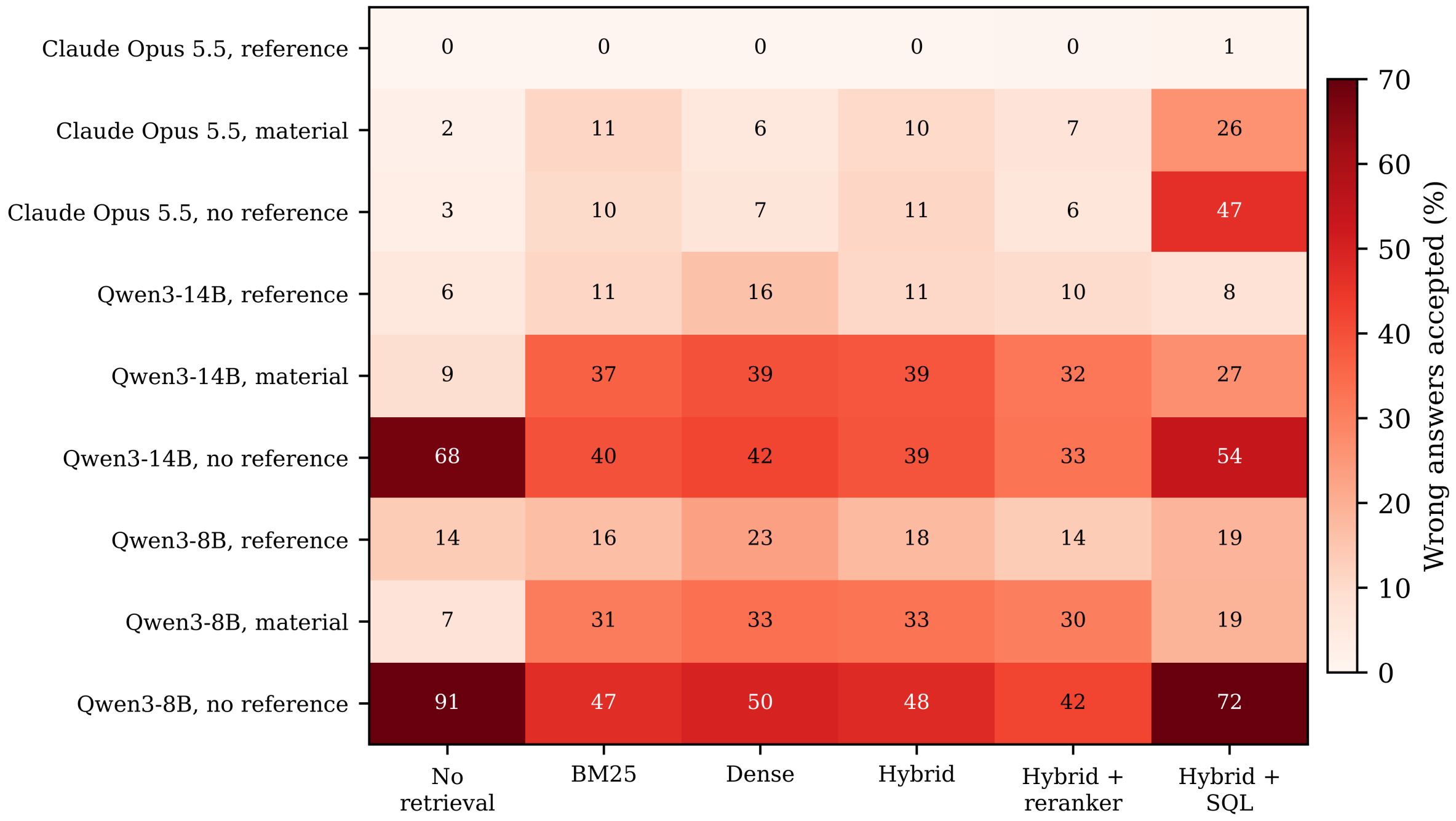

| | No retrieval | BM25 | Dense | Hybrid | Hybrid + reranker | Hybrid + SQL |
|---|---|---|---|---|---|---|
| Claude Opus 5.5, reference | 0 | 0 | 0 | 0 | 0 | 1 |
| Claude Opus 5.5, material | 2 | 11 | 6 | 10 | 7 | 26 |
| Claude Opus 5.5, no reference | 3 | 10 | 7 | 11 | 6 | 47 |
| Qwen3-14B, reference | 6 | 11 | 16 | 11 | 10 | 8 |
| Qwen3-14B, material | 9 | 37 | 39 | 39 | 32 | 27 |
| Qwen3-14B, no reference | 68 | 40 | 42 | 39 | 33 | 54 |
| Qwen3-8B, reference | 14 | 16 | 23 | 18 | 14 | 19 |
| Qwen3-8B, material | 7 | 31 | 33 | 33 | 30 | 19 |
| Qwen3-8B, no reference | 91 | 47 | 50 | 48 | 42 | 72 |

Figure 4: **Wrong answers accepted across retrieval systems.** Heatmap displaying the percentage of incorrect generations falsely accepted by each judge across the six architectures (values printed within each cell). For each judge, the three rows represent evaluation with the reference answer (top), retrieved material (middle), and question and answer only (bottom).

### 3.4 Error Taxonomy Across Inquiry Types and Domain Ambiguities

Stratifying evaluation fidelity across task categories highlights the intrinsic structural limits of reference-free grading. For pairwise material comparisons and longitudinal trends, reference-free Claude Opus verdicts perform no better than random chance ($\varkappa \leq 0.09$), whereas reference augmentation restores robust agreement ($\varkappa = 0.90$ and $0.92$, respectively). Because trend and comparative inquiries require evaluating shifts between cohort medians spanning hundreds of independent publications, neither candidate prose nor retrieved text snippets provide sufficient evidentiary scope for evaluation.

Crucially, evaluative breakdown is not confined to literature-scale statistical synthesis; it pervades single-paper lookups—a task conventionally assumed to be straightforward for retrieval-augmented generation. When grading single-paper parameter lookups without a reference key, Claude Opus exhibits high false acceptance rates, accepting 30.0% of incorrect answers under evidence passages and 35.0% unassisted. Because physical science papers report extensive matrices of control cells, unoptimized baselines, and secondary device architectures alongside champion devices, candidate models frequently extract plausible but non-champion metrics. An unanchored judge inspecting a fluent natural-language excerpt cannot determine whether a reported 16.5% efficiency represents the paper's champion single-junction cell, an unpassivated control, or a secondary device. Furthermore, evaluation audits identify domain-specific ambiguities that cannot be resolved without explicit ground-truth rules: whether peak device metrics encompass multijunction tandems or non-standard illumination protocols, whether binary cation designations permit minor ternary additives, and whether a claim that no literature records qualify is factual or represents an extraction omission under stringent query thresholds. The indoor-light ambiguity is also present in the source data: without protocol metadata, a judge cannot distinguish standard one-sun illumination from indoor measurements. An audit of the device records identified 13 human-curated records from eight papers, at least seven of them from papers on indoor or dim-light cells, whose efficiencies exceed the single-junction record of their year by more than 1.5 percentage points, and six of the 755 computed answers rest on them. Excluding these six questions leaves every $\tau$ unchanged and changes $\varkappa$ by at most 0.005, confirming that benchmark conclusions remain robust.

# 4 Conclusion

This work provides an empirical assessment of language-model evaluators on quantitative scientific literature tasks by benchmarking 4,530 free-text generations against deterministic ground truth computed from 45,441 experimental photovoltaic records. Our results demonstrate that model-based evaluation in the physical sciences succeeds only when anchored by computed reference keys and precise tolerance rules: Claude Opus reproduces true benchmark scores with near-perfect fidelity ($\varkappa = 0.982 \pm 0.003$) and preserves exact architectural rankings ($\tau = 1.00$). Withholding reference ground truth triggers severe evaluation breakdown: frontier agreement declines to $\varkappa = 0.557$, evidence-passage augmentation provides no measurable relief ($\varkappa = 0.576$), and local open-weight judges invert system rankings to place hallucinating closed-book models first ($\tau \leq 0.20$). Moreover, structured SQL-derived outputs systematically mislead evaluators without a reference: the frontier judge accepts 46.6% of wrong SQL answers, and the local judges 53.9% and 72.3%.

These findings establish an operational guideline for scientific AI evaluation: automated natural-language verification cannot replace structured empirical data. In quantitative domains, evaluating literature-scale synthesis requires ground truth precomputed from relational measurements, establishing systematic database curation as a prerequisite not only for generating scientific answers, but for validating the systems and evaluators that produce them.

**Data availability.** The benchmark questions, computed answers, generated answers and judge verdicts are not publicly available. The curated device records derive from the Perovskite Database [9], and the retrieved passages come from full texts whose licences do not permit redistribution.

**Acknowledgements.** The authors used large language models as assistants: Claude (Anthropic) for drafting parts of the manuscript and developing the analysis code, and Gemini (Google) for help with the writing. Claude (Anthropic) and the open Qwen3 models are also the experimental subjects evaluated in this study. All text, code, and empirical results were verified, curated, and revised by the authors, who retain full responsibility for the manuscript.

**Competing interests.** The authors declare no competing interests.